\documentclass[a4paper,UKenglish]{tgdk}

\usepackage{graphicx}
\usepackage{paralist}

\title{Large Language Models and Knowledge Graphs: Opportunities and Challenges}
\titlerunning{LLMs and KGs: Opportunities and Challenges}

\author{Jeff Z. Pan}{The University of Edinburgh, United Kingdom   \and \url{https://knowledge-representation.org/j.z.pan/} }{j.z.pan@ed.ac.uk}{https://orcid.org/0000-0002-9779-2088}{}

\author{Simon Razniewski}{Bosch Center for AI, Germany   }{Simon.Razniewski@de.bosch.com}{}{}

\author{Jan-Christoph Kalo}{Vrije Universiteit Amsterdam, The Netherlands  }{j.c.kalo@vu.nl}{}{}

\author{Sneha Singhania}{Max Planck Institute for Informatics, Germany  }{ssinghan@mpi-inf.mpg.de}{}{}

\author{Jiaoyan Chen}{The University of Manchester \& University of Oxford, United Kingdom }{jiaoyan4ai@gmail.com}{}{}

\author{Stefan Dietze }{GESIS - Leibniz Institute for the Social Sciences \& Heinrich-Heine-Universität Düsseldorf, Germany  }{stefan.dietze@hhu.de}{}{}

\author{Hajira Jabeen}{GESIS - Leibniz Institute for the Social Sciences, Germany  }{hajira.jabeen@gesis.org}{}{}

\author{Janna Omeliyanenko }{University of Würzburg, Germany }{omeliyanenko@informatik.uni-wuerzburg.de }{}{}

\author{Wen Zhang}{Zhejiang University, China  }{zhang.wen@zju.edu.cn}{}{}

\author{Matteo Lissandrini }{Aalborg University, Denmark  }{matteo@cs.aau.dkn}{}{}

\author{Russa Biswas }{Hasso-Plattner Institute, Germany }{russa.biswas@hpi.de}{}{}

\author{Gerard de Melo }{Hasso-Plattner Institute, Germany }{gerard.demelo@hpi.de}{}{}

\author{Angela Bonifati }{Lyon 1 University, France }{angela.bonifati@univ-lyon1.fr}{}{}

\author{Edlira Vakaj }{Birmingham City University, United Kingdom }{edlira.vakaj@bcu.ac.uk}{}{}

\author{Mauro Dragoni }{Fondazione Bruno Kessler, Italy }{dragoni@fbk.eu}{}{}

\author{Damien Graux}{Edinburgh Research Centre, CSI, Huawei Technologies UK,  United Kingdom}{damien.graux@huawei.com}{}{}

\authorrunning{Jeff Z. Pan et al. }

\Copyright{Jeff Z. Pan, Simon Razniewski, Jan-Christoph Kalo, Sneha Singhania, Jiaoyan Chen, Stefan Dietze, Hajira Jabeen, Wen Zhang, Matteo Lissandrini, Russa Biswas, Gerard de Melo, Angela Bonifati, Edlira Vakaj, Mauro Dragoni and Damien Graux}

\subjclass{, 
}


\DOI{10.1234/0000000.00000000}
\Volume{000}
\Issue{111}
\Article{42}
\DateSubmission{Date of submission} 
\DateAcceptance{Date of acceptance} 
\DatePublished{Date of publishing} 
\AssociateEditor{Editor Name}
\nolinenumbers

\date{April 2023}

\begin{document}

\maketitle

\begin{abstract}

Large Language Models (LLMs) have taken Knowledge Representation---and the world---by storm. This inflection point marks a shift from explicit knowledge representation to a renewed focus on the hybrid representation of both explicit knowledge and parametric knowledge.  In this position paper, we will discuss some of the common debate points within the community on LLMs (parametric knowledge) and Knowledge Graphs (explicit knowledge) and speculate on opportunities and visions that the renewed focus brings, as well as related research topics and challenges. 

\end{abstract}


\section{Introduction 
}






Large Language Models (LLMs) have taken Knowledge Representation (KR)---and the world---by storm, as they have demonstrated human-level performance on a vast spectrum of natural language tasks, including some tasks requiring human knowledge. Following this, people are gradually   starting to accept the possibility of having knowledge represented in the parameters of some language models. The arrival of LLMs announces the era of Knowledge Computing, in which the notion of reasoning within KR is broadened to many computation tasks based on various knowledge representations.

This is a big step for the field of Knowledge Representation. For a long time, people focused on explicit knowledge, such as those embedded in texts, sometimes also known as unstructured data, and those in a structured form, such as in databases and knowledge graphs (KGs)~\cite{Pan2017}. Historically, for a long time, humans used texts to pass down their knowledge from one generation to another, until around the 1960s, when researchers started to study knowledge representation for better natural language understanding and developed early systems, such as ELIZA~\cite{Weiz1996} at the MIT. In the early 2000s, the Knowledge Representation and the Semantic Web communities worked together to standardize the widely used knowledge representation languages, such as RDF~\cite{Pan2009} and OWL~\cite{GHMP+2008}, at web scale, using which the large-scale knowledge bases are then more widely known as KGs~\cite{Pan2017}, due to their helpful graph structures, enabling the both logical reasoning and graph-based learning.

This inflection point, with the arrival of LLMs, marks a paradigm shift from explicit knowledge representation to a renewed focus on the hybrid representation of both explicit knowledge and parametric knowledge. As a popular approach for explicit knowledge representation, KGs are now widely investigated for the combination with Transformer-based LLMs, including pre-trained masked language models (PLMs) like BERT \cite{devlin2018bert} and RoBERTa \cite{liu2019roberta}, and more recent generative LLMs like the GPT series \cite{brown2020language} and LLaMA \cite{touvron2023llama}.
Some works use LLMs to augment KGs for, e.g., knowledge extraction, KG construction, and refinement, while others use KGs to augment LLMs for, e.g., training and prompt learning, or knowledge augmentation. In this paper,  considering both directions, LLMs for KGs and KGs for LLMs, we present a better understanding of the shift from explicit knowledge representation to a renewed focus on the hybrid representation of both explicit knowledge and parametric knowledge.



A related survey paper~\cite{zhu2023llms} presents a comprehensive review of using LLMs for KG construction and reasoning, while our work provides a more in-depth view of the inflection point, considering not only relational KGs but also KGs with ontologies as schemas, as well as other dimensions of structured knowledge, including tabular data~\cite{XCKL+2018} and numerical values~\cite{PaHo2003}. 
%
Other works on the intersection of LLMs and KGs have a minor overlap with the topics covered in our paper; e.g., on studies using LLMs as KGs~\cite{alkhamissi2022review},  on using KGs to augment LLMs~\cite{yang2023chatgpt}, or on comparing GPT-4 with ChatGPT and  SOTA fine-tuning methods on three knowledge-related tasks---entity, relation and event extraction, link prediction, and KG question answering~\cite{zhu2023llms}. Overall, none of these papers look into the implications of the inflection point with concrete applications. To this end, this paper summarises the common debate points within the community, introduces the state-of-the-art for a comprehensive set of topics where KGs and LLMs are integrated, and further presents opportunities and challenges.

\section{Common Debate Points within the Community}\label{sec:debate-points}

The usage of 
parametric and explicit knowledge together is a topic of debate in the Knowledge Computing community, with proponents and skeptics offering different perspectives. Below are some summaries of common points of contention.

\textbf{Knowledge Representation and Reasoning:}
KGs offer a structured representation of knowledge with explicit relationships, enabling reasoning and inference~\cite{MePa08}. Critics argue that parametric knowledge in LLMs relies on statistical patterns rather than true understanding and reasoning~\cite{Bender2020ClimbingTN}. Proponents of LLMs like ChatGPT, highlight their ability to generalize from large-scale text corpora, capturing a range of information, and excellent language understanding capabilities. On the one hand, LLMs could generate plausible but incorrect or nonsensical responses, such as hallucinations, due to a lack of explicit knowledge representation \cite{Zhang2023ACS}. There are also doubts on whether LLMs have the ability to learn directional entailments~\cite{LHWS2022} or infer subsumption between concepts~\cite{he2023language}.
On the other hand, KGs can be costly to build.
While LLMs can be expensive to train too, they can be readily usable to support many downstream applications, bringing AI from the backstage to centre stage.  Thus parametric knowledge is not the (only) destination for LLMs. To sum up, in comparison to the classic trade-off between expressiveness and decidability in Knowledge Representation, here we have the trade-off between precision and recall considering using explicit and parametric knowledge in Knowledge Computing tasks.

\textbf{High Precision Methods:} The success of KGs can largely be attributed to their ability to provide factual information about entities with high accuracy. For instance, YAGO~\cite{Suchanek2007YagoAC} asserts an accuracy rate exceeding 95\%. Similarly, Google necessitates high accuracy in its KG for operational use, e.g., the semi-automatical construction method of  Knowledge Vault was not utilized in production, partly due to its inability to reach the desired 99\% accuracy in their benchmark \cite{weikum-machine-knowledge}. Along this line of thought, many LLM-based methodologies for KG completion fail to reach these high-performance levels, as exemplified by the performance of BERT in~\cite{LHPVP2022,wdknown}, and GPT-3, equipped with hundreds of billions of parameters~\cite{alivanistos2022prompting}. This calls for novel high precision methods for KG construction based on LLMs.

\textbf{Numerical Values:} It is widely recognized that LLMs grapple with handling numerical values. 
Even straightforward arithmetic tasks can be a struggle for LMs, as highlighted in a study by Big-bench~\cite{srivastava2022beyond}. This shortcoming also extends to KG completion tasks~\cite{kalo2022kamel}. Multiple LLMs have been evaluated on their ability to complete KGs using numerical facts from Wikidata~\cite{wikidata}, such as individuals' birth and death years. However, none of the tested models accurately predicted even a single year. This raises questions about the capability of current LLMs to correctly memorize numbers during pre-training in a way that enables them for subsequent use in KG completion. While LLMs like PaLM~\cite{srivastava2022beyond} demonstrate some proficiency in dealing with numbers, more commonly used smaller models seem ill-equipped for this task. The complexity escalates when considering the intricacies of metrics and diverse numbering formats and types. Currently, modifying LLMs to handle numerical values remains unresolved, making their utilization for numerical KG completion seem far from practical.

\textbf{Long-tail Knowledge:} 
One of the key research questions on LLMs for the Knowledge Computing community (and beyond) is how much knowledge LLMs remember~\cite{mallen2023trust}. 
Investigations indicate that LLMs' performance significantly deteriorates when dealing with random Wikidata facts, specifically those associated with long-tail entities, in comparison to popular entities, as evidenced in the PopQA dataset~\cite{mallen2023trust} and other datasets~\cite{petroniLAMA2019,wdknown}. 
This effect can be traced back to a causal relationship between the frequency of an entity's appearance in the pre-training corpus and the LLMs' capacity for memorization~\cite{elazar2023measuring}. Even sizable LLMs face difficulties when trying to retain information about long-tail entities~\cite{kandpal2022large}. KGs inherently present an advantage over LLMs through their provision of knowledge about long-tail entities~\cite{kalo2022kamel,wdknown} and thus can further help improve the recall for Knowledge Computing tasks.

\textbf{Bias,  Fairness and Beyond:} Critics argue that LLMs can perpetuate and amplify biases present in the training data, leading to biased outputs. LLMs may generate biased or prejudiced responses if the training data contains stereotypes or discriminatory information \cite{Nadeem2020StereoSetMS, Leidinger2023WhichSA}. On the other hand, proponents argue that bias is not inherent to LLMs but reflects societal biases embedded in the data. They emphasize the importance of addressing bias in the training data and developing mitigation techniques \cite{Sap2019TheRO, Plank2022TheO, Ranaldi2023ATT}. 
A survey~\cite{Blodgett2020LanguageI} argued that ``bias'' in Natural Language Processing (NLP) systems took various conceptualizations without being critically engaged by practitioners. 
KGs are also used in a plethora of downstream tasks, and social biases engraved in KG embeddings get propagated~\cite{Groth2022KnowledgeGA, Kraft2022TheLO}. Specifically, ontology creation, which generally comprises manual rules factored by opinions, motivations, and personal choices, is a source of bias~\cite{Janowicz2018DebiasingKG, Du2022UnderstandingGB}. Also, automated pipelines for KG construction exhibit gender bias \cite{Mehrabi2019ManIT}. There are other similar concerns of LLMs beyond bias and fairness, including (but not limited to) copyright violation and misinformation. In general, due to the implicit nature of parametric knowledge, it is less straight forward to forget such toxic information from LLMs, compared to explicit knowledge. 


\textbf{Explainability and Interpretability:}
KGs are often preferred in scenarios where explainability and interpretability are crucial~\cite{CLPHC2018}, as they explicitly represent relationships between entities and provide a structured knowledge representation. Skeptics of LLMs argue that these models lack transparency and interpretability, making it difficult to understand how they arrive at their answers or recommendations. Proponents of LLMs acknowledge the challenge of explainability but argue that recent research efforts~\cite{Arrieta2019ExplainableAI, Jain2019AttentionIN} are improving LLM's interpretability through techniques like attention mechanisms, model introspection.
Some also argue that Chain-of-Thoughts (CoT)~\cite{Wei2022ChainOT} can also improve the explainability of LLMs, although question decomposition and precisely answering sub-questions with LLMs are still far from being solved. 
Attribution evaluation and augmentation of LLMs with e.g., source paragraphs and sentences is another recent research topic for improving their explainability in question answering \cite{bohnet2022attributed}.

\section{Opportunities and Visions 
}

One of the key questions this paper needs to answer is, now with the emergence of parametric knowledge, what new opportunities do we have? Here are some of our thoughts on such new opportunities with the arrival of parametric knowledge and its potential integration with explicit knowledge. 


\begin{enumerate}
    \item\textbf{Instant access to huge text corpora:} As mentioned in the Introduction, for a long time, human beings passed down their knowledge in texts. Thus, a lot of knowledge these days are in textual form.  Using LLMs gives access to extremely large text corpora at high speed, and recently even on consumer hardware \cite{hu2021lora}. 
    This allows AI developers to avoid getting bogged down in previously critical challenges around data gathering, preparation, storage, and querying at scale. It also helps to reduce previously critical dependencies on the field of information retrieval.
 \item\textbf{Richer knowledge for many subtasks:} Although the most prominent capabilities of LLMs, question answering and dialogue, are still under critical scrutiny, it should not be overlooked that LLMs have significantly advanced and simplified many traditional tasks of the knowledge engineering pipeline. Out-of-the-box, with fine-tuning on a few examples, or via few-shot prompting, LLMs have advanced many tasks such as dependency and structured parsing, entity recognition, and relation extraction. And just as errors propagate along a pipeline, so do improvements, thus enabling KG construction at unprecedented scale and quality.
Furthermore, LLMs are readily usable for many downstream tasks beyond knowledge engineering. By injecting explicit, and in particular structured, knowledge into LLMs, such as through retrieval augmented methods, one can make explicit knowledge more readily usable for such wide range of downstream tasks, further realising the vision of `Knowledge 
is power'.

 \item\textbf{Even more advanced language understanding:} LLMs alone already significantly advanced the ``understanding'' of natural language, as evidenced by tasks like textual entailment, summarization, paraphrase detection and generation, etc. These capabilities are critical to making knowledge engineering robust to linguistic variance, typographic errors, redundancy, and other features of human-written, web-scraped, and other noisy forms of text. Now with potential novel approaches to combining parametric knowledge with explicit knowledge, it is possible to have even more advanced language understanding, not only for textual entailments, but also for other NLP tasks, such as summarization and consistent generation.  

 \item\textbf{Compression entails consolidation:} An important step in traditional knowledge engineering is the consolidation and aggregation of conflicting and concurring pieces of information, requiring often elaborate methods for consolidating observations from sentences, patterns, and constraints \cite{shin2015incremental}. In LLM training, an aggregation occurs automatically. Although this step is not entirely understood, it brings the potential for outsourcing a major challenge in knowledge engineering.
\end{enumerate}

\noindent
With the above new opportunities brought by the combination of parametric and explicit knowledge, 
our vision is two-folded:
\begin{itemize}
    \item In \textbf{Explicit-Knowledge-First} use cases, our vision is that \textit{LLMs will enable, advance, and simplify crucial steps in the knowledge engineering pipeline so much as to enable KGs at unprecedented scale, quality, and utility}.
    \item In \textbf{Parametric-Knowledge-First} use cases, our vision is that \textit{KGs will improve, ground, and verify LLM generations so as to significantly increase reliability and trust in LLM usage}.
\end{itemize}

\noindent
Both visions are neither alternatives, nor does one build upon the other. Instead, we believe that classes of use cases will continue to exist side-by-side, some of which favor   Explicit-Knowledge-First (scrutable) approaches, some of which favor Parametric-Knowledge-First (blackbox) approaches, with either of these having significant potential for benefiting from synergies of the two directions.

\section{Key Research Topics and Related Challenges}

With the opportunities and visions related to the availability of both parametric and explicit knowledge  in place, in this section, 
we categorize, summarize, and present the recent developments in using LLMs and KGs under four different themes.

\subsection{LLMs for KGs: Knowledge Extraction and Canonicalisation}

KG construction is a complex task that demands collecting and integrating information from a wide array of sources, encompassing structured, semi-structured, and unstructured data. Traditional methods often rely on specific modules designed to process each data type in isolation and struggle when the content is diverse and heterogeneous in structure. However, LLMs are powerful NLP models trained on a broad spectrum of information sources, making them well-suited for knowledge extraction tasks. This section presents work that uses LLMs for knowledge extraction from various sources.

\subsubsection{Entity Resolution and Matching 
}\label{sec:entity-linking}

Entity resolution (also known as entity matching, entity linking or entity alignment) is the process of linking pieces of information occurring in multiple heterogeneous datasets and referring to the same world entity \cite{Elmagarmid07,fellegi69,2021Papadakis}. 
Past research has focused on developing the methodologies and similarity measures among entities represented by flat structured data.  However, entity resolution for semi-structured data for KGs is a fairly recent topic with significantly less attention. Approaches for entity alignment can be split into general vs embedding-based categories.

\textbf{General approaches}, such as CG-MuAlign \cite{Zhu2020} which employs Graph Neural Networks (GNNs) to perform multi-type entity alignment, leverages the neighborhood information and generalizes to unlabeled types, and REA~\cite{Pei20} which tackles the multi-lingual entity alignment problem by combining adversarial training with GNNs to cope with the problem of noisy labeled data provided as input. 

\textbf{Embedding-based} entity alignment methods for KGs reduces the symbolic similarities between graph entities to a vector space in order to flatten the heterogeneity of graph components and facilitate reasoning~\cite{SunZHWCAL20}. Specifically, a total of 23 representative embedding alignment approaches are cross-compared in terms of performance but also are shown to require significant supervision in the labeling phase.  Therefore, unsupervised methods and methods that cope with large-scale KGs are highly desirable in future research investigations. 

LLMs are used in entity resolution and linking for KGs in multiple ways \cite{abs-2306-01388}.
First, LLMs can help with labeling training data, which is typically a resource-intensive and time-consuming step, hindering the performance of entity alignment for KGs.
Similar to how \cite{Shao0WR20} employs Generative Adversarial Networks (GANs) to reduce the effort of labeling data, we argue that LLMs can provide labeled samples of KGs 
and control the performances of the aforementioned embedding-based approaches.  
Also, LLMs can help build solid corpora of entity matching rules, modulo the fact that a declarative formalism with a logical language $L$ is defined in the graph setting. Training data for this logical language should be provided as input to LLMs, similarly to SQL statements that are readily available for consumption in text corpora. However, prompt engineering is needed in order to produce meaningful rule corpora for real-world large-scale KGs, such as DBpedia~\cite{auer2007dbpedia} and Wikidata~\cite{wikidata}. Entity matching rule logs can be envisioned for these real-world large-scale KGs in a similar fashion as query logs for these KGs \cite{BonifatiMT19,BonifatiMT20}. 

Concluding, entity alignment and matching are necessary pre-processing steps for full-fledged knowledge reasoning. The combination of general entity linking approaches with embedding-based ones, as well as the leveraging of LLM-driven rule and labeled data construction, can lead to better integration of LLMs with knowledge reasoning \cite{hu2022empowering}. The latter integration of LLMs and knowledge reasoning can also boost performance, thus enabling the interpretability and explainability of the model's output and filling the gap between symbolic and statistical AI.

\subsubsection{Knowledge Extraction from Tabular Data 
}
Extracting knowledge from tabular data like databases, Web tables and CSV files is a common way for KG construction.
For tables whose semantics (meta information) are already known, heuristic rules can be defined and used to transform their data into KG facts.
However, real-world tables often have unclear semantics with important meta information, such as table name and column header, not clearly defined. Meanwhile, the original data usually need to be retrieved, explored, integrated and curated, before expected knowledge can be extracted. 
In recent years, Transformer-based LMs have been investigated for processing tables especially their textual contents. 
They can be applied to table vector representation as a foundation of other prediction tasks \cite{vogel2023towards}. 
TURL~\cite{deng2022turl} is a typical method of table representation learning that uses BERT~\cite{devlin2018bert} and has been applied in several tasks such as cell filling, column type annotation, and relation extraction.
Similarly, RPT~\cite{tang2020rpt} uses BERT and GPT to pre-train a table representation model. 
Starmie~\cite{fan2022semantics} transforms columns into sequences using a template and fine-tunes BERT with a contrast learning framework using unionable and not unionable column pairs as samples.

Among all the table processing tasks, semantic table annotation which matches table data to KG components (e.g., table column to KG class, table cell to KG entity, inter-column relationship to KG property) can be directly applied to extract knowledge for KG construction and population \cite{liu2022tabular,jimenez2020semtab}.
There have been several attempts that use LLMs for these tasks.
Doduo \cite{suhara2022annotating} serializes a table into a sequence of tokens and trains BERT for predicting column types and inter-column relationships.
Korini et al. \cite{korini2023column} prompts ChatGPT to annotate semantic column types. When task-specific samples for demonstration are minimal or absent, ChatGPT achieves similar performance RoBERTa model. Although some attention has been given to utilizing LLMs for tabular data processing and KG construction, there is still much room for investigation, especially towards the following challenges: 
\begin{itemize}
    \item \textbf{Transforming table contents into sequences:} A table or a table element with its structured context needs to be transformed into a sequence before it can be fed into LLMs. Different transformation methods are required for different LLM utilization scenarios, such as fine-tuning LLMs, LLM inference with prompts, and instruction tuning of LLMs.
    \item \textbf{Representing and utilizing none textual tabular data:} A table often contains not only long and short text but also data of other types like numbers and dates. There are still few works that consider these data. 
    \item \textbf{Extracting tabular knowledge:} LLMs are mostly applied to process and understand tables but rarely applied to the final step of knowledge extraction. OntoGPT \cite{caufield2023structured}, which extracts instances from texts to populate an ontology using ChatGPT is known, but there are no counterparts for tables. Beyond instances, extracting relational facts is more challenging.
\end{itemize}

\subsubsection{Knowledge Extraction from Text 
}

Knowledge extraction from text generally entails automatic extraction of entities and associated relations, with traditional pipelines processing vast amounts of sentences and documents. This process enables the transformation of raw text into actionable knowledge, facilitating various applications such as information retrieval, recommendation systems, and KG construction. The language understanding capabilities of LLMs have enhanced this process.  

For example,
\begin{inparaenum}[(1)]
\item Named Entity Recognition (NER) and Entity Linking, as mentioned in Section~\ref{sec:entity-linking}, involve identifying and categorizing named entities (such as persons, organizations, and locations) in text and linking (more in Section~\ref{sec:linking}) them to KGs. 
\item Relation extraction focuses on identifying and classifying relationships between entities, with LLMs that leverage zero-shot and few-shot in-context learning techniques~\cite{Wei2023ZeroShotIE, Li2023EvaluatingCI}. 
\item Event extraction aims to detect and classify events mentioned in the text, including their participants and attributes~\cite{wadden-etal-2019-entity, Zhang2023COFFEEAC}.
\item Semantic Role Labeling (SRL) involves identifying the roles played by entities in a sentence, such as the subject, object and predicate~\cite{Shi2019SimpleBM, zhong-chen-2021-frustratingly}.
\end{inparaenum}

These approaches allow LLMs to extract information from text without needing a large amount of explicit training on specific domains, thereby improving their versatility and adaptability. Moreover, LLMs have demonstrated proficiency in extracting knowledge from languages other than English, including low-resource languages, paving the way for cross-lingual knowledge extraction and enabling the utilization of LLMs in diverse linguistic contexts~\cite{Lai2023ChatGPTBE}.

Furthermore, prompting LLMs has introduced new paradigms and possibilities in the field of NLP. LLMs can generate high-quality synthetic data, which can then be used to fine-tune smaller task-specific models. This approach, known as synthetic data generation, addresses the challenge of limited training data availability and enhances the performance of models~\cite{Josifoski2023ExploitingAF, Tang2023DoesSD}. Additionally, instruction tuning has emerged as a powerful technique where LLMs are trained on datasets described by explicit instructions, enabling more precise control and customization of their behavior for specific tasks~\cite{Wei2023ZeroShotIE, Wang2023InstructUIEMI}. Also, for constructing domain-specific KGs, the stakes are higher, and hence scrutinizing the generated text (by experts) is necessary. However, it is still a step forward since human annotation is less expensive than human text generation.

Apart from the obvious substantial computational resource requirements for training and utilizing these LLM, there are various challenges, including the ones presented in Section~\ref{sec:debate-points}. More specifically, the following future directions are still possible:
\begin{itemize}
    \item \textbf{Efficient extraction from very long documents.} Current LLMs are incapable of processing very long documents like novels in one shot. In this regard, modeling long-range dependencies and performing corpus-level information extraction can be further improved.
    \item \textbf{High-coverage information extraction.} Almost all extraction pipelines focus on high precision. However, high recall is ignored or under-explored~\cite{10.1162/tacl_a_00456}. Building knowledge extractors with high precision and high recall will result in a great leap toward building lifelong information extractors.
\end{itemize}


\subsection{LLMs for KGs: Knowledge Graph Construction 
}

We highlight the important role that LLMs play in improving KG construction, focusing on current trends, issues, and unanswered questions in this field.
We start by discussing link prediction, which is a way to generate new facts given an existing KG.
Next, we look at inductive link prediction, a method that predicts triples for unseen relationships. 
Our focus then shifts to a more recent method where triples is directly extracted from the parametric knowledge of an LLM.

As a conclusion of this section, we discuss the challenges of LLM-based methods for KG construction. 
These involve issues with long-tail entities, numerical values, and also the precision of these methods.

\subsubsection{Link Prediction 
}\label{sec:linking} 
Link prediction is about predicting a missing element of a triple given the other two elements.
It includes head entity prediction $(?, r, t)$, relation prediction $(h,?,t)$, and tail prediction $(h, r, ?)$. 


KG link prediction methods have mostly been studied for static snapshots of a KG.
Many approaches, in fact, assume a training phase in which the current state of the KG is used to learn embeddings and other model parameters.
Subsequently, many such models can only operate on entities for which an embedding was learned in the training phase.
This leaves them incapable of predicting links for any previously unseen entities, such as newly added people or products.
Inductive link prediction (ILP), in contrast, focuses on techniques that can predict links to new entities not originally contained in a KG. 
Furthermore, existing KG embedding-based KG completion approaches frequently fail to leverage textual information and other literal information~\cite{gesese2021survey}.

To overcome these challenges, current research focuses on incorporating textual information available in the KGs to improve the KG embeddings, boost performance in downstream tasks, and to support ILP.
The latent representation is learned from textual information using a variety of encoding models such as linear models, convolutional models, recurrent neural models, and LLMs and studied in~\cite{lu2020utilizing,pan2023unifying}. 
In this work, we focus only on LLM-based methods.

The LLM encoder BERT~\cite{devlin2018bert} is used in Pretrain-KGE~\cite{zhang2020pretrain} to generate initial entity embeddings from entity descriptions and relations. 
These embeddings are then fed into KG embedding models to generate final embeddings. MADLINK~\cite{biswas2021madlink} uses SBERT to generate embeddings from entity descriptions, together with the entity embeddings obtained from structural information. KEPLER~\cite{wangKEPLERUnifiedModel2021} offers a unified approach for KG embedding and pre-trained language representation, embedding text-enhanced knowledge and factual knowledge into LLMs. Nayyeri et al.~\cite{nayyeri2022integrating} use LLMs to produce representations at word, sentence, and document levels, merging them with graph structure embeddings. Huang et al.~\cite{huang2022endowing} propose a model that combines LLMs with other modalities, such as vision, to generate a multi-model embedding space. CoDEx~\cite{alam2022language} uses a novel loss function driven by LLMs that helps KG embedding models estimate the likelihood of triples based on textual information.
While these approaches can leverage the structural information contained in the graph, KG embeddings can not be directly used to predict unknown entities in the ILP setting. To still predict entities within ILP, existing works that combine text embeddings and traditional KG embeddings use only the text embeddings in the ILP setting \cite{malaviya2020commonsense} or apply similarity based heuristics to generate KG embeddings for unseen entities \cite{commonkg}.

Instead of considering the graph structure, another line of research directly leverages LLMs for KG Completion.
For example, KG-BERT~\cite{yao2019kgbert} represents a triple as 
an ordered sequence of head text, including surface form, descriptions, and attributes, relation text, tail text separated with special tokens. 
KG-BERT~\cite{yao2019kgbert} optimizes the BERT model on KGs, followed by KG-GPT2~\cite{biswas2021contextual} that fine-tunes GPT-2 model. MTL-KGC~\cite{kim-etal-2020-mtl-kgc} enhances the effectiveness of KG-BERT by combining prediction and relevance ranking tasks. PKGC evaluates triple validity by translating it into natural language sentences, while LLMs process these sentences for binary classification. Masked Language Model (MLM) is introduced to encode KG text, with MEMKGC~\cite{choi2021mem}
predicting masked entities using the MEM classification model. Open world KGC~\cite{choi2023knowledge} expands MEMKGC with modules Entity Description Prediction (EDP) and Incomplete Triple Prediction (ITP), focusing on predicting entities with given textual descriptions. 
StAR~\cite{wang2021structure} uses Siamese-style textual encoders for text and a scoring module, while SimKGC~\cite{wangSimKGCSimpleContrastive2022a} leverages a Siamese textual encoder. LP-BERT~\cite{li2022lp} is a hybrid KG completion method that combines MLM encoding for pre-training with LLM and separated encoding for fine-tuning, using a contrastive learning strategy.
%
%
Also, LLMs such as GPT-3 and GPT-4 have the in-context learning capability which could adapt to new tasks through careful prompt design without fine-tuning the model parameters. 
Since LLMs are directly capable of predicting entities that are not yet contained in the KG, many works using direct LLM approaches also evaluate their models in the ILP setting \cite{choi2021mem,choi2023knowledge,dazaBLP2021,wang2021structure,wangSimKGCSimpleContrastive2022a,wangKEPLERUnifiedModel2021,zha2022rulesinductive}.

Beyond the described approaches that construct and leverage embeddings for link prediction, LLMs may also be directly used in a prompting setting to find suitable links between entities, with existing methods described in detail in the following Section \ref{topics:kgconstruction:extraction}. 
As with the previous approaches that purely use LLMs, prompt-based approaches are directly applicable for ILP and are commonly evaluated in this setting \cite{bosselut-etal-2019-comet,jiang2020can,petroniLAMA2019,shinAutoPrompt2020,zhong2021factual}.

{\textbf{Challenges and Opportunities:}} 
LLMs make it significantly easier to jointly utilize structural and text information for link prediction, while there are still challenges:
\begin{itemize}
    \item The generative language model which uses a decoder-only or encoder-decoder architecture cannot ensure that the generated result is already included in the KG. Also, one entity may have multiple natural language names. Thus it is hard to judge the correctness of the generated results from LLMs.
    \item Current link prediction models are mostly evaluated on ranking metrics, such as Hit@$k$ and  Mean Reciprocal Rank, which requires methods to give a ranked list of candidates. Considering that the number of candidate entities might be huge (e.g., over ten thousand), framing each candidate as an sample requires too many times of predictions and LLMs can not be efficiently evaluated due to its large scales (e.g., over 100 billion parameters) and high computation cost. 
    \item The key research question of link prediction is how well a method could learn to infer new triples based on existing ones. LLMs are trained based on a massive corpus that might overlap with KGs such as Wikidata~\cite{wikidata}. 
    Thus it is not easy to distinguish whether the LLM completes the prediction by utilizing its memory 
    or reasoning over existing facts. 
    \item LLMs provide several benefits for ILP as demonstrated by the many existing approaches, but their application also has certain drawbacks. 
    Since ILP entities may occur that are not contained within the underlying KG, the task is inherently reliant on auxiliary information. 
    When leveraging LLMs for ILP, many studies focus on improving the extraction of knowledge from LLMs through prompt engineering, which is, by itself, a current popular research area that may directly benefit the domain.
    \item While prompting provides promising results, it requires well-designed strategies for predicting multi-token entities and out-of-vocabulary tokens.
    Further, if required knowledge is not captured by the LLM, e.g., when querying novel concepts that emerged after LLM pre-training, schemes to incorporate further knowledge into the LLM are required. Meanwhile, as feature selection, finding a suitable prompt also needs much searching with many experiments, which is costly especially for those recent commercial LLMs like GPT-4.
    \item Alternative methods rely on available high-quality textual descriptions of unknown entities. 
    These methods inherit the limitations of classical link prediction, in that they require one-vs-all comparisons against all entity candidates during inference, which may become computationally prohibitive for large KGs. 
    As such, efficient strategies for obtaining predictions may provide a promising research direction.  
    \item Further, the combination of these methods with classical link prediction models has already shown strong performance and may be further explored to incorporate structural information into LLM-driven approaches.
\end{itemize}
Though challenges exist, opportunities lie in designing efficient and effective link prediction methods combining LLMs preserving the efficiency of traditional methods and robustness of language models.

\subsubsection{LLMs for KGs: Triple Extraction from LLMs} 
\label{topics:kgconstruction:extraction}
Traditionally, retrieval and reasoning of relational knowledge have both relied on symbolic knowledge bases 
{\cite{10.1007/978-3-319-25007-6_28}}, that often are constructed using supervised extraction techniques applied to unstructured corpora, e.g. Web archives \cite{journals/semweb/YuGFLRD19,tempelmeier2018inferring}. More recently, self-supervised LLMs have been investigated for their ability to directly retrieve relational knowledge~\cite{zhou2023large} from their parameters, e.g. through question answering, prompting through the use of cloze-style questions~\cite{heinzerling-inui-2021-language,sachan-etal-2021-syntax} or statement scoring~\cite{tamborrino2020pre}. 
In this context, the ability of LLMs to retrieve, infer and generalize relational knowledge is seen as a crucial indicator of their capacity to understand and interpret natural language. Even though a range of terms are used in that context, e.g. fact or knowledge retrieval as well as knowledge inference, we refer to the task of accessing relational knowledge from LLM parameters as \textit{knowledge retrieval}.

\textbf{Benchmarks and Baselines:} 
LAMA is the first benchmark dataset 
to evaluate knowledge retrieval in LLMs \cite{petroni-etal-2019-language}.
Related works show that knowledge retrieval through prompts is inconsistent with regard to paraphrasing \cite{elazar2021measuring, heinzerling-inui-2021-language}, with some types of information guiding LLMs towards more correct answers \cite{cao-etal-2021-knowledgeable, petroni2020context, chen2022knowprompt}, while other
are harmful to their performance \cite{pandia2021sorting, kassner-schutze-2020-negated}. LLMs struggle to retrieve knowledge from low-frequency phenomena \cite{ravichander2020systematicity} and \cite{Hwang2021COMETATOMIC2020} argue that LLMs fail to express large varieties of knowledge when prompted for it in a zero-shot manner. 

Zhong et al. \cite{zhong2021factual} propose that the models' accuracy may be from memorizing training data, not actually inferring knowledge. 
Similar to LAMA, the experiments on a more recent probing work KAMEL~\cite{kalo2022kamel} confirm that LLMs are still far from the knowledge access capabilities of symbolic knowledge bases. The Knowledge Memorization, Identification and Reasoning test work KMIR~\cite{gao2022kmir} reveals that while LLMs struggle to robustly recall facts, their capacity to retain information is determined more by the number of parameters than the training methods, and while model compression can help preserve the memorization performance, it reduces the ability to identify and reason about the information in LLMs from transformer-based language models. Linzbach et al. \cite{linzbach-etal-prompt-syntax-nlp4kgc-2023} also present similar findings.
LLMs are known to struggle with more complex reasoning tasks  \cite{huang2022towards,he2023language}. Branco et al. \cite{branco-etal-2021-shortcutted} explore generalisability of common-sense reasoning capabilities and the impact of shortcuts in training data. 

\textbf{Biases in Triple Extraction Evaluation:} LLMs may exhibit various types of biases; representation of the majority viewpoint being a common issue due to distributions prevalent within pretraining data~\cite{bender2021dangers}, neglecting disagreements among multiple viewpoints (e.g. by majority voting) ~\cite{davani2022dealing}. Prior works investigate individual factors (such as frequency) or LLM biases in other tasks~\cite{Mao-9881877}, as well as knowledge retrieval~\cite{zhong2021factual}. 
 
 With respect to the interpretation, reliability and generalisability of knowledge retrieval, several studies~\cite{branco-etal-2021-shortcutted,cao-etal-2021-knowledgeable} investigate whether LLMs actually learn transferable generalisations or only exploit incidental shortcuts in the data. 
\cite{cao-etal-2021-knowledgeable} explore biases in three different knowledge retrieval paradigms, namely \textit{prompt-based retrieval}, \textit{case-based analogy}, \textit{context-based inference}, finding that decent performance of existing knowledge retrieval baselines tends to be driven by biased prompts that overfit to artefacts in the data, guide the LLM towards correct entity types or unintentionally leak correct answers or additional constraints applicable to the correct answer.  
In a similar context,~\cite{du2022shortcut} discuss the shortcut learning behaviour arising due to skewed training datasets, the model, or the fine-tuning process. \cite{schramowski2022large} demonstrate an intriguing similarity between human cognitive biases and those exhibited by LLMs. Using insights from psychology, they analyse the learning and decision-making processes of black-box models to reveal their biases towards right-and-wrong for decision-making.
Therefore, rigorous assessment of existing benchmark datasets is necessary for generalizable insights about knowledge retrieval and inference performance, and to facilitate efficient, unbiased knowledge retrieval from LLMs.

\textbf{Prompt Engineering for Triple Extraction:}
Cao et al. \cite{cao-etal-2021-knowledgeable} propose three paradigms for factual knowledge extraction from LLMs: prompt-based, case-based, and context-based. Results suggest the prompt-based retrieval is biased towards prompt structure. Prompt engineering~\cite{bach2022promptsource} aims to create prompts that efficiently elicit desired responses from LLMs for a specific task.
However, a limited number of manually created prompts only reveal a portion of the model's encoded knowledge~\cite{jiang2020can}, as the response can be influenced by the phrasing of the question. Thus, prompt engineering is a crucial part of knowledge retrieval from LLMs.
LPAQA~\cite{jiang2020can} uses an automated mining-based and paraphrasing-based method to 
generate diverse high-quality prompts, as well as ensemble methods to combine answers from different prompts.
Automatic Prompt Engineer~\cite{zhou2023large} uses LLM models like InstructGPT~\cite{ouyang2022training} and instruction induction~\cite{honovich2022instruction} to generate instruction candidates which are then improved by proposing semantically similar instruction variants to achieve human-level performance.
Zhou et al. \cite{zhou2023large} investigate the ability of LLMs, such as GPT-3, to generate high-quality prompts for a variety of tasks.
Initial experiments on the role of syntax in knowledge retrieval~\cite{linzbach-etal-prompt-syntax-nlp4kgc-2023} find a strong dependency on prompt structure and knowledge retrieval performance.

To summarise, prior works have shown that relational knowledge is captured by LLMs to a certain extent. However, there is still insufficient understanding of how performance differs across different kinds of knowledge or relations, for instance, commonsense knowledge compared to entity-centric encyclopedic facts or transversal versus hierarchical relations. In addition, several studies raise questions about LLMs capacity to infer knowledge beyond already-seen statements.

\textbf{Challenges and Opportunities: }
\begin{itemize}
    \item \textbf{Entity Disambiguation:} Entity disambiguation is essential for KG construction to ensure unique identification of entities and to maintain consistency across the graph. However, when extracting facts from LLMs, entity disambiguation presents several challenges, since LLMs only operate on word token level. Hence, polysemy and homonymy make it difficult to determine the correct entity when a term has multiple meanings or is spelled the same as others but has different meanings. Also, the need to resolve co-references, where the same entity is mentioned in various ways within a text, further complicates the process. Moreover, the same piece of text can refer to different entities depending on the context, making it a significant challenge to correctly identify and classify the entities. 
    Entities that were not present in the training data or are less common in general can be particularly hard to disambiguate. This can be a frequent issue with newer, less well-known, or very domain-specific entities.
    These complexities lead to major challenges that need to be addressed: enhancing disambiguation techniques to better handle long-tail entities; developing methods to better understand and utilize context in entity disambiguation; and improving co-reference resolution in a way that it can be effectively incorporated into KG construction.  
    \item \textbf{Long-tail Entities:} Existing LLMs still manifest a low level of precision on long-tail entities. Models may begin to generate incorrect information when they fail to memorize the right facts. The answers provided by these models often lack consistency. Incorrect correlations drawn from the pre-training corpus can lead to various biases in KG completion.
    Whether retrieval-augmented models serve as a viable solution to this problem remains uncertain, as does the potential necessity to adapt pre-training and fine-tuning processes to enhance model robustness in handling long-tail entities.
    \item \textbf{High-Precision:} LLMs face challenges in achieving high-precision predictions when performing knowledge extraction~\cite{wdknown}.
    A potential strategy to derive high-precision KGs from LLMs is to focus on model calibration.
    However, there are pressing challenges that remain unsolved. How can LLM training be adapted to prioritize high-precision learning? Can LLMs be used for validation purposes? These questions form the crux of the ongoing exploration in this field.
    \item \textbf{Provenance:} Extracting factual knowledge directly from LLMs does not provide provenance, the origin and credibility of the information, which presents multiple issues. 
    Without provenance, verifying the accuracy of information becomes challenging, potentially leading to the spread of misinformation. 
    Additionally, bias detection is hindered, as the lack of source information makes it difficult to account for potential biases in the data used for training. Provenance also provides critical context, without which, information can be misunderstood or misapplied. Lastly, the absence of source information compromises model transparency, making it hard to evaluate the accountability of the LLMs.
\end{itemize}

\subsection{LLMs for KGs: Ontological Schema Construction 
}
A KG is often equipped with an ontological schema (including rules, constraints and ontologies) for ensuring quality, enabling easier knowledge access, supporting reasoning, etc. Meanwhile, an independent ontology, which usually represents conceptual knowledge sometimes with logics, can also be regarded as a KG. In this part, we introduce topics that LLMs are applied to learn ontological schemas and to manage ontologies.

\subsubsection{Constraint and Rule Mining from KGs 
}


The existing KGs are mostly obtained from large-scale data extraction pipelines, which are notoriously brittle and can introduce errors and inconsistencies in these graphs~\cite{Dong:KFusion14,rabbani:Extraction23}.
Moreover, a KG is \emph{never considered complete} since the closed world assumption does not hold~\cite{Dong:KFusion14,Paulheim:KGRefinement17},
i.e., it is not possible to conclude that a missing fact is false \emph{unless it contradicts another existing fact}.
Instead, we usually consider that in a KG it holds the open-world assumption, that is a missing fact is simply considered as \emph{unknown}.
%
%

Practical applications impose high demands in terms of (semi-)automatic methods for data quality assessment and validation~\cite{knublauch:shapesSHACL17,rabbani2022shaclposter,RuleMining:Ahmadi20}.
Since KGs contain huge amounts of data, 
it is not feasible to manually inspect and correct their errors.
Therefore, a common approach is to instantiate rules and constraints that can be automatically enforced. 
These constraints express dependencies and conditions that the KG needs to satisfy at all times and that should not be violated by the introduction of new facts or their deletion.
In KGs, rules and constraints can take the form of Graph Functional Dependencies~\cite{Fan:GraphFun20}, declarative first-order logic rules~\cite{Galarraga:AMIE13}, or \emph{validating shapes}~\cite{knublauch:shapesSHACL17,prud2014shape}.
Once a set of rules or constraints are instantiated, the next step is to either identify which entities or facts in the KG violate any of them, or employ them to delete erroneous information, or, finally, to employ them to deduce any missing information~\cite{fan2019deducing,Rabbani:SHACTOR23}.

\begin{example}\label{ex:rules}
The following rules could apply for a subset of a graph describing people and their relationships:

$$
r_1: \mathtt{hasParent}(x, y)\Rightarrow\mathtt{hasChild}(y, x);
$$
$$
r_2: \mathtt{hasParent}(x, y)\Rightarrow{\exists}{v_1,v_2{\in}\mathbb{N}}~|~\mathtt{hasBirthYear}(x, v_1)\wedge \mathtt{hasBirthYear}(y, v_2);
$$
$$
r_3: \mathtt{hasChild}(x, y){\wedge}\mathtt{hasBirthYear}(x, v_1){\wedge}\mathtt{hasBirthYear}(y, v_2)\Rightarrow {v_1}<{v_2};
$$
Where $r_1$ states that \texttt{hasChild} is the inverse equivalent relation of \texttt{hasParent}, $r_2$ states that for each person in the KGs for which we know the parent-child relationship we should know the birth year, and $r_3$ states that if $y$ is a child of $x$ then $x$ should be born before $y$.
\end{example}

Nonetheless, a fundamental challenge is how to generate such rules and constraints.
Specifying them manually is prohibitively difficult and expensive~\cite{RuleMining:Ahmadi20,rabbani2022shaclposter}.
On the one hand, the domain experts, who know the semantics for the dataset at hand, may not have the skill set or the background necessary to formally express those rules. 
Even when skilled, domain experts would require a substantial amount of manual work to exhaustively materialize a complete list of such rules \cite{rabbani:Extraction23}.
Therefore, in the past decade, we have witnessed an increased interest in methods that can (semi-) automatically extract rules and constrains~\cite{rabbani:Extraction23,zhang2022knowledge}.
In all those cases, the data in the KG is exploited as a \emph{silver standard}~\cite{Paulheim:KGRefinement17}.
That is, while the data are not perfect, we can still assume it is at least of reasonable quality.
This approach works for rules that can find frequent positive examples.
The challenge arises for rules that need to govern entities and facts that are rare, and for negative rules that need to prohibit some facts from appearing.
To handle these cases, contextual information can be exploited. In this context, LLMs bring a number of important opportunities: 

\begin{itemize}
\item ability to extract contextual information from input text;
\item information extraction from contextual data used during training;
\item automatic generation of new rules via inductive reasoning;
\item understanding vocabulary in context, allowing to reconcile rules with synonyms and polysemy;
\item providing explanations of rules or generating candidate validating or counterfactual examples.
\end{itemize}


Consider, for example, a rule where the relationships \texttt{hasParent} has been used, an LLM can be interrogated to identify whether  \texttt{hasParent} between two entities refers more likely to the semantic meaning of being biologically descended or being owned or controlled.
LLMs can be used to filter out candidate constraints, e.g., given their ability to mimic simple common sense reasoning, rules can be translated in the form of questions to the LLM, e.g., ``\textit{if X is child of Y is it true that the birth year of Y comes always before the birth year of X?}''.
Thus, an LLM can be used in support or replacement of a human annotator towards generated rules.

One of the most promising abilities of LLMs is parsing long texts.
In companies and organizations, documents exist that contain reference governing information, e.g., procedures, regulations, and specifications.
Here we see an untapped opportunity in parsing these documents in relation to the entities and predicates in the KG to extract constraints.
Yet, the challenge arises in the fact that the LLM needs to use the correct vocabulary of entities and relations and the correct rule syntax. 
Recent advancements in program synthesis show a possible way to the challenge.

Finally, an even more fundamental challenge is that of transcending the usage of LLMs for NLP alone, and using them directly on large sets of facts within a KG.
We can think of this setting as a special kind of multi-modal LLMs, where the KG is a specific modality.
This brings to the question of whether it is possible to train LLMs to treat the task of rule generation as, for example, a summarization task.
This would require then the ability to perform both inductive and abductive reasoning and treat rules as summaries from the set of facts in the KG.

\noindent\textbf{A Rule Mining Case with ChatGPT} Let's have a look at the following real example asking ChatGPT to generate rules. 
\begin{example}
We ask ChatGPT ``Please list three rules that could help to infer one person's father.  And output it in a first-order rule form.'' The following is the output we get:

\begin{quotation}
``Certainly! Here are three rules expressed in first-order logic form that could help infer a person's father:

Rule 1: $\forall x, y \; (\texttt{child}(x, y) \land \texttt{male}(y) \to \texttt{father}(y, x))$
This rule states that for all individuals x and y, if there exists a 'child' relationship between x and y and y is male, then y is the father of x.

Rule 2: $\forall x, y \;  (\texttt{biologicalChild}(x, y) \to \texttt{father}(y, x))$
This rule states that for all individuals x and y, if there exists a 'biologicalChild' relationship between x and y, then y is the father of x.

Rule 3: $\forall x, y \; (\texttt{fatherOf}(y, x) \to \texttt{father}(y, x))$
This rule states that for all individuals x and y, if there exists a 'fatherOf' relationship between y and x, then y is the father of x.

These rules, expressed in first-order logic, demonstrate how the relationships between individuals can be used to infer the father of a person within a KG.''
\end{quotation}
\end{example}
From the above example, we can see that ChatGPT gives three rules that can possibly infer the father relationship.
Rule 1 is correct which is consistent with the definition of father; Rule 2 is a rule with high probability; and Rule 3 is also correct since $\texttt{father}$ and $\texttt{fatherOf}$ are inverse relations.
This demonstrates LLMs have rule knowledge and could express it in a formal form following instruction, thus they could serve as experts for automatic rule construction. 
To this end, however, we need to \textbf{align the rules from the LLM to the existing knowledge in the KG}, and \textbf{feed the LLM about the existing knowledge and the context of the KG}. Both are consistent with discussion about the company and family examples.


\subsubsection{Ontology Refinement 
}

Ontology refinement includes quite a few topics like knowledge completion (e.g., subsumption prediction, complex concept learning and new concept placement), erroneous knowledge detection and repair (e.g., inconsistency checking) and knowledge canonicalization (e.g., entity renaming).
Besides formally represented knowledge, real-world ontologies, such as the widely used medical ontology SNOMED CT\footnote{\url{https://www.snomed.org/}} and food ontology FoodOn\footnote{\url{https://foodon.org/}}, also include a lot of meta information defined by different annotation properties for usability, such as entity labels, synonyms and natural language definition.
Taking the concept \textit{obo:FOODON\_00002809} in FoodOn as an example, it has not only formal knowledge such as named super concepts and logical restrictions, but also labels and synonyms (e.g., ``edamame''), definitions (e.g., ``Edamame is a preparation of immature soybean ...''), comments and so on. 
These meta information, especially the natural language text, further motivates people to use LLMs for ontology refinement.

For a refinement task, usually there are quite a few existing examples in the original ontology.
Therefore, a straightforward solution, which has been adopted by most current methods, is fine-tuning a Pre-trained Language Model such as BERT together with an attached classifier.
One typical method is BERTSubs \cite{chen2023contextual} which is to predict the subsumption relationship between two named concepts, or between one named concept and one complex concept. 
It concatenates the corresponding texts of the two candidate concepts with special tokens as the input of a model composed of a pre-trained BERT and a logistic regression classifier, and fine-tunes the model with the existing subsumptions in the target ontology.
For a named concept, the text could be either its name (label or synonym), or its name in combination with a textual description of its surrounding concepts; while for a complex concept, the text is its description (a.k.a. verbalisation). 
Another typical work is \cite{liu2020concept} which fine-tunes BERT and predicts the position to place in SNOMED CT for a new concept. 
Note that there are also some language model-based methods in taxonomy curation, such as \cite{shi2023subsumption} which fine-tunes BERT for taxonomy edge completion and GenTaxo \cite{zeng2021enhancing} which fine-tunes a BERT variant named SciBERT for predicting positions that need new concepts.  
They can be directly applied or easily extended to refine an ontology's concept hierarchies.  

Exploiting LLMs is a promising direction for ontology refinement, but it still needs much effort before they become practical tools. DeepOnto \cite{he2023deeponto}, which is a Python-based package that can support quite a few ontology engineering tasks, has already included some tools for ontology refinement and alignment using LLMs, but more development is needed to make it more accessible and to support generative LLMs like LLaMA and GPT-4.
One obvious challenge is that those very recent generative LLMs have been rarely explored for ontology engineering. However, we think the following two research challenges are more fundamental:
\begin{itemize}
  \item \textbf{Exploiting the graph structure and logics of an ontology together with its text}. Currently LLM fine-tuning-based methods can well utilize the text of individual entities, but their other formal semantics are often ignored or not effectively incorporated. Besides fine-tuning with samples constructed by some templates, more LLM techniques such as prompt learning and instruction tuning could be considered.  
  \item \textbf{Combing symbolic reasoning with LLM inference}. Symbolic reasoning, such as consistency checking in OWL ontologies, can still play a role to e.g., validate the knowledge inferred by LLMs. One aspect to incorporate symbolic reasoning is constructing samples for LLM fine-tuning and extracting prompts for LLM inference, while another aspect is the synergized framework \cite{zhu2023llms} where LLM inference and symbolic reasoning work iteratively.
\end{itemize}

\subsubsection{Ontology Alignment 
}
The content of one single ontology is often incomplete and many real-world applications rely on cross-domain knowledge. 
Ontology alignment (a.k.a. ontology matching), which is to identify cross-ontology mappings between entities that have an equivalent, subsumption or membership relationship, thus becomes especially important for knowledge integration.
The entity can be a concept (class), an individual (instance) or a property.
Traditional systems (e.g., LogMap \cite{jimenez2011logmap}) heavily rely on lexical matching and optionally use symbolic reasoning to remove mappings that lead to logical conflicts; while some recent methods combine these techniques with machine learning techniques like feature engineering, semantic embedding and distant supervision for better performance (e.g., LogMap-ML \cite{chen2021augmenting}). 
Especially, when the ontologies have a large ratio of assertions (large ABoxes) and the task is to discover equivalent individuals, ontology alignment is very close to the KG entity alignment task that has been widely investigated in recent years using KG embeddings \cite{zhang2020industry}. 
As in ontology refinement, exploiting the textual information by applying LLMs is a promising direction for augmenting ontology alignment.

The study of LLM application in ontology alignment is similar to ontology refinement.
Pre-trained language models such as BERT have been applied via fine-tuning \cite{neutel2021towards,he2022bertmap}.
BERTMap \cite{he2022bertmap} is a typical system that has achieved state-of-the-art performance in many biomedical ontology alignment benchmarks.
It fine-tunes a pre-trained LM with synonym pairs extracted from the original ontologies and the potentially given mappings, and combines the predicted concept equivalence scores with lexical matching scores and reasoning for the mappings.
Those recent LLMs like GPT-4 have not been applied in ontology alignment, as far as we know, and the two fundamental research challenges mentioned in ontology refinement are applicable in ontology alignment. 
Besides, ontology alignment has another critical challenge: 
\begin{itemize}
\item
\textbf{Evaluating LLM-based ontology alignment systems}.
Novel evaluation protocols with new metrics are needed to fairly and efficiently compare LLM-based systems even with incomplete ground truth mappings \cite{he2022machine}. Meanwhile, the semantics from the textual meta information and the LLM may be inconsistent with formal semantics defined in ontologies, and thus it is sometimes hard to determine whether a mapping by an LLM-based system is true or not.

\end{itemize}

\subsection{KGs for LLMs: Training and Accessing LLMs}


In Sections 4.1 to 4.3, we discussed on three different aspects on using LLMs for KGs. In this section, we look into the other direction, i.e., using KGs for LLMs. There are a few dimensions here. Firstly, KGs can be used as training data for LLMs. Secondly, triples in KGs can be used for prompt construction. Last but not least, KGs can be used as external knowledge in retrieval augmented language models.

\subsubsection{KGs for (Pre-)Training Language Models 
}

KGs usually contain information extracted from highly trusted sources, post-processed, and vetted by human evaluations. Information from KGs has been integrated into the pre-training corpus since natural language text alone can lead to limited information coverage \cite{yao2019kgbert, peters2019knowledge, agarwal2020knowledge, Xu2023KILMKI}. 

Using factual knowledge from KGs to pre-train LLMs has also infused structured knowledge \cite{Moiseev2022SKILLSK}. This integration of KGs with LLMs, along with efficient prompts, has made it convenient to inject world knowledge and incorporate new evolving information into language models \cite{Santos2022KnowledgePI}. Additionally, knowledge expressed in high-resource language KBs has been transferred into LMs tuned for low-resource languages \cite{Zhou2021PrixLMPF, Liu2021LearningDR}.

Furthermore, grounding knowledge from KGs to pre-train LMs has shown improvements in performance on generation and QA tasks \cite{Chen2020KGPTKP, Ribeiro2020InvestigatingPL, Ouz2020UniKQAUR}. In another approach, \cite{verga2020facts} proposed an interpretable neuro-symbolic KB, where the memory consists of vector representations of entities and relations from an existing KB. These representations are augmented to an LM during pre-training and fine-tuning, enabling the model to excel in knowledge-intensive QA tasks.

\subsubsection{KGs for Prompt Construction  }\label{sec:kgs-prompt-construction}

The attention received by the integration of KGs and LLMs has grown recently.
On the one hand, there is the explored direction of prompting LLMs for collecting and distilling knowledge in order to make it available to the end-users.
On the other hand, there is the less explored research direction where KGs are used in synergy with prompts in order to enhance LLMs with capabilities making them more effective and, at the same time, trustworthy.
A number of studies have leveraged KGs to enrich and fine-tune prompt creation resulting in a significant increase in prompt quantity, quality, and diversity compared to manual approaches. 
KGs have been employed in single and in multi-turn scaffolding prompts at scale, powered by numerous traversal paths over KGs with low authoring cost while considering the meaningful learning patterns~\cite{10.1145/3491102.3502087}. 
Other studies have investigated how incorporating explicit knowledge from external sources like KGs can help prompt engineering, especially by giving additional contexts (e.g., attributes, K-hop neighbors) of the entities in order to help the LLMs to generate better predictions~\cite{brate2022improving}.
Approaches like KnowPrompt~\cite{chen2022knowprompt} use KGs to incorporate semantic and prior knowledge among relation labels into prompt-tuning for relation extraction, enhancing the prompt construction process and optimizing their representation with structured constraints.
Certain studies have utilized LLMs and prompts in the task of reasoning over KGs~\cite{choudhary2023complex}, e.g., LARK uses entities and relations in queries to find pertinent sub-graph contexts within abstract KGs, and then, performs chain reasoning over these contexts using LLM prompts of decomposed logical queries outperforming previous state-of-the-art approaches by a significant margin.\\

\noindent
\textbf{Challenges and Opportunities:} The current research in the field of KG utilization for prompt creation predominantly centers around LLMs, which are considered to have relatively lower efficacy compared to LLMs. 
LLMs present significant potential for advancing prompt creation methodologies in conjunction with KGs. We may summarize this perspective within the following four challenges:
\begin{itemize}
\item \textbf{C1:} KGs can be employed to automatically extract and represent relevant knowledge to generate context-aware writing prompts. Analyze and understand the relationships between different writing prompts, enabling the generation of prompts that build upon each other. 
\item \textbf{C2:} KGs can be combined with LLMs to facilitate the interactive and dynamic generation of prompts, adapting to user feedback and preferences in real time. Furthermore, the use of KGs in prompt creation has opened up possibilities for explainability and interpretability. 
Since KGs provide a transparent representation of knowledge, the prompts generated from KGs can be easily traced back to their underlying sources.
\item \textbf{C3:} KGs can integrate into prompts the definitions of guards exploited during the generative task. Such guards may lead to enhancing the trustworthiness of the information generated by LLMs and make them more compliant with specific domain-wise or context-wise constraints.
\item \textbf{C4:} KGs can create prompts that ask questions (e.g., inferring missing relations in an incomplete KG)  that trigger KG complex reasoning capabilities and
intermediate reasoning steps.
\end{itemize}


The integration of KGs within the prompt construction activities will allow us to answer the following preparatory set of research questions.
For each research question, we provide a link to the challenges mentioned above, aiming to identify appropriate research pathways.

\begin{itemize}
\item \textbf{RQ1:} How can KGs be integrated into existing prompts to enhance the effectiveness (including relevance, non-biased, and privacy-preserving) of the information extracted from LLMs? $\longrightarrow$ C1, C2, C3.
\item \textbf{RQ2:} How can KGs be exploited to drive the generative capabilities of LLMs in order to properly address the whole ethical constraints of AI-based solutions? $\longrightarrow$ C2, C3.
\item \textbf{RQ3:} What are the optimal approaches to generate KGs-based prompts that enhance reasoning capabilities? $\longrightarrow$ C4.
\end{itemize} 



Also, in the following three scenarios, the community would benefit from tackling the above-mentioned challenges and research questions.

\begin{enumerate}
\item \textbf{KGs for Hallucination Detection in LLMs:}
The reliability of LLMs is greatly affected by the hallucination problem, where they generate inaccurate information. 
Despite attempts to address it, the issue of hallucination is likely to persist in the realm of LLMs for the foreseeable future.
To aid in the detection of hallucinations, KGs-based prompting aims to offer reliable information that can serve as a foundation.
By combining LLMs and KGs, researchers can develop a comprehensive prompt-based fact-checking model that can identify hallucinations in various domains.

\item \textbf{KGs for Editing Knowledge in LLMs:}
LLMs possess the ability to store extensive real-world knowledge, but they struggle to exploit prompts to update their internal knowledge to reflect real-world changes. 
KGs-based prompts offer a potential solution for modifying knowledge in LLMs, but they are restricted to handling basic tuple-based knowledge in KGs. 
Indeed, even if the entire LLM undergoes re-training, the knowledge presented through prompts would likely be assimilated within the vast network structure of the LLM.

\item \textbf{KGs for Black-box LLMs Knowledge Injection:}
While pre-training and knowledge editing methods can bring LLMs up to date with the latest information, they require access to the internal structures and parameters of LLMs. 
However, many cutting-edge large LLMs only offer APIs that allow users and developers to interact with them, keeping their internal workings hidden from the public. 
Consequently, traditional KG injection techniques that involve modifying LLM structures with additional knowledge fusion modules cannot be employed. 
One potential solution is to convert various types of knowledge into different text prompts. 
However, it remains an area of ongoing research to determine if these prompts can effectively adapt to new LLMs. 
Additionally, the approach of using KGs-based prompts is constrained by the length of input tokens accepted by LLMs. 
Therefore, the question of how to enable effective knowledge injection for black-box LLMs still remains unanswered.
\end{enumerate}

\subsubsection{Retrieval Augmented Methods 
}

There are a few of reasons that retrieval augmented methods are necessary for  LLMs to obtain external knowledge. One reason   
  is to address the problem of knowledge cutoff, i.e., LLMs are not aware of   the events that happened after their training. Also, although parametric knowledge would increase when the size of parameters increases, training LLMs is expensive; e.g.,  GPT-3 (175B parameters) costs \$4.6 million to train, and PaLM (540B parameters) costs \$17 million. In fact, research suggests that the obtained knowledge from such training is mainly about popular entities~\cite{mallen2023trust}. Furthermore, for domain specific applications, there might be some significant knowledge that is not yet in LLMs, including private and business critical knowledge that cannot be put into LLMs. 

  One idea to deal with the above lack of (updated) knowledge is to edit the knowledge in LLMs. A  obvious strategy is to retrain and fine-tune the model based on the modified data. 
However, apart from being costly,   retraining cannot guarantee that erroneous data will be corrected. Another Strategy  is to  develop hyper-network to learn a parameter shift for the base model. De Cao et al.
\cite{de-cao-etal-2021-editing} trained a hyper-network, KnowledgeEditor, to modify a fact and used Kullback-Leibler (KL) divergence-constrained optimization to alleviate the side effect on other data/knowledge that should not be changed. However, this method does not perform well when editing \textit{multiple edits}, as it uses the same strategy to process multiple edits and ignore the relation between different edit gradients, resulting in a ``zero-sum'' phenomenon, where the inter-gradient conflict will inevitably cause some data modifications to fail. Han et al.~\cite{HLLP2023}   design explicit and implicit multi-editor models to learn diverse editing strategies in terms of dynamic structure and dynamic parameters respectively,  allowing to deal with the conflict data in an efficient end-to-end manner. 
 
However, the above Knowledge Editing methods are not yet scalable, people started to introduce  retrieve-generate architectures for building retrieval augmented generation models. These methods are mainly using unstructured passages as external knowledge. RAG \cite{NEURIPS2020_6b493230} outperforms DPR~\cite{KOML+2020} by marginalizing the retrieval step to train the generator and retriever jointly with the supervision of the label answer. FiD \cite{IzGr2021} encodes the concatenation of the passages retrieved by pre-trained DPR  and the original question separately, and then fuses them with concatenation to the decoder. 
It is expected that structured knowledge will be the main source of external knowledge, as passages often contain noise. Knowledge Graphs can be used directly as external knowledge. They can also be used to enhance passage-based methods~\cite{YZFY+2022}. 

Retrieval augmentation is a very promising direction. There are a few pressing challenges:

\begin{itemize}
    \item {\bf Unifying Knowledge Editing and Retrieval Augmentation}: KGs can be used for editing knowledge in LLMs, while at the same time, KGs can also be used as external knowledge to assist LLMs in retrieval augmented methods. In fact, knowledge editing and retrieval augmentation is getting very close. For example, Mitchel et al. \cite{mitchell2022memory} proposed a Retrieval-Augmented Counterfactual Model (SERAC), which stores edits in an explicit memory for knowledge editing over LLMs. 

     \item {\bf Semi-parametric LLMs}:  This direction is highly related to the topic of this position paper. The idea is to make use of explicit knowledge to augment LLMs. One of the key issue  is to integrate different explicit knowledge~\cite{TLRJ+2023}, including unstructured ones, such as passages, and structured ones, such as KGs and databases, for augmenting LLMs. 
  
      \item {\bf Support of Complex Reasoning}: Can we go beyond simply retrieving explicit knowledge by enabling reasoning through retrieval augmented methods?   BehnamGhader et al.~\cite{BMR2022}  demonstrated with their experimental results that the similarity metric used by the retrievers is generally insufficient for reasoning tasks. Furthermore, LLMs do not take the complicated relations between   statements into account, thus leading to to poor reasoning performance.

     
\end{itemize}

\subsection{Applications }

The integration of KGs and LLMs in a unified approach holds significant potential, as their combination mutually enhances and complements each other in a valuable manner. For instance, KGs provide very accurate and explicit knowledge, which is crucial for some applications i.e. healthcare, whereas LLMs have
been criticized for their lack of factual knowledge leading to hallucinations and inaccurate facts. secondly, LLMs lack explainability instead, KGs given their symbolic
reasoning ability, are able to generate interpretable results. On the other hand, KGs are difficult to construct from unstructured text and suffer from incompleteness therefore, LLMs could be utilized in addressing these challenges by text processing. 
Various applications have adopted this methodology of combining LLMs with KGs, such as  healthcare assistants\footnote{https://neo4j.com/blog/doctor-ai-a-voice-chatbot-for-healthcare-powered-by-neo4j-and-aws/}, question answering systems \cite{yasunaga2021qa} or ChatBots, and sustainability, among others.

\subsubsection{Commonsense Knowledge 
} 

The majority of KGs capture facts of the sort one might encounter in an encyclopedia or in a relational database.
However, commonsense knowledge is another important form of world knowledge for AI systems. For instance, we may wish for a KG to not only capture that the Congo rainforest lies in Central Africa, but also that tropical rainforests have significant rainfall and lush green vegetation. ConceptNet is the most well-known commonsense knowledge graph, developed using manual crowdsourcing along with automated refinement techniques \cite{ConceptNet}.
However, crowdsourcing is very labor-intensive and costly, so alternative means of harvesting such knowledge have long been sought.

\textbf{Commonsense Knowledge from LLMs:} The first study to investigate extracting knowledge from a language model to the best of our knowledge was indeed one that targeted commonsense knowledge \cite{TandonDeMelo2010}. The authors mined commonsense triples such as \texttt{hasProperty (apples, green)} from the Google Web 1T n-gram data as well as from Microsoft's 
Web-scale smoothed language models \cite{MicrosoftWebLM}. This was later extended into a large-scale commonsense knowledge graph \cite{TandonDeMeloWeikum2011} that covered a range of different relations and became a part of the WebChild KG~\cite{Tandon2014WebChildHA}.

As both crowdsourcing and information extraction from text are likely to lead to incomplete knowledge, a key challenge is how to generalize beyond what has been collected.
The WebBrain project explored neural knowledge graph completion \cite{WebBrain2016} for better generalization. 
COMET, short for COMmonsEnse Transformer \cite{bosselut-etal-2019-comet}, and the improved COMET-ATOMIC 2020 \cite{Hwang2021COMETATOMIC2020}, used existing data to fine-tune Transformer-based models.
This line of work considers the original ConceptNet relations as well as reasoning-related knowledge pertaining to events, causes, and effects, e.g., what goals might have motivated 
a person $A$ to leave an event without person $B$.
Recently,~\cite{west2021symbolic} shows how common-sense triples could be extracted from an LLM and use through distillation to transfer knowledge into a smaller LM, outperfoming the larger one.
Overall, fine-tuned LLMs are found to outperform off-the-shelf LLMs, while also benefitting from the advances of the latter.

\textbf{Challenges and Opportunities:} Commonsense knowledge, in particular, is genuinely open-ended, such that it depends on a number of considerations whether it makes sense to attempt to materialize relevant knowledge beforehand or rather invoke a (possibly slow) LLM on the fly.
Commonsense knowledge may also differ substantially between different cultures \cite{Nguyen2023}. This also leads to the question of what kinds of biases are acceptable.
Finally, a long-term challenge is how to capture knowledge that is not easily expressed in language, e.g., how a robot ought to grasp different kinds of objects.

\subsubsection{Digital Build Environment 
}

In the domain of Build Environment, where it is vital to design and construct in a safe and sustainable way, a number of regulations and guidelines need to be met. Automated Compliance Code Checking has bloomed to support this, but still, in the past, it was quite challenging to interpret regulations and execute rules in 3D models automatically due to the ambiguity of the text, and the need for extensive expert knowledge for interpretation. Many applications now in this area have combined LLMs + KGs in addressing these challenges by following an LLMs enhanced KG approach \cite{zhu2023llms}. LLMs interpret the text in the regulations and enhance a KG of rules, which are further serialized using dome Domain Specific Language \cite{zhang2023transformer, wang2023deep}. Interrogating 3D models modeled as graphs using the Linked Building Data approach is another challenge for the domain as it requires skill sets in query languages like SPARQL. LLMs are helping in understanding human language written questions and converting those into relevant query languages by bringing a new way of how domain experts interact and interrogate 3D models and their various forms. An example of this is AI Speckle \footnote{https://speckle.systems/blog/ai-powered-conversations-with-speckle/}.

\subsubsection{Digital Healthcare 
}

The Digital Healthcare sector holds immense potential for various possibilities concerning the adoption of LLMs, including the automation of clinical documentation, the synthesis of patient histories, and the identification of potential candidates for clinical trials.
Although these advancements are remarkable, it is crucial to recognize the potential risks associated with employing LLMs in healthcare.
Indeed, Digital Healthcare is one of the most critical application domains for the adoption of LLMs.
The needs of the major stakeholders (i.e., physicians, healthcare providers, and policymakers) row against the paradigm behind the creation of LLMs.
In particular, the two major significant risks related to the model's accuracy and the privacy concerns stemming from its usage.

\textbf{Accuracy}.
Some demonstrations of LLMs have showcased impressive capabilities. 
Nevertheless, there have also been documented instances where LLMs have made mistakes or exhibited erratic behavior. 
In the Digital Healthcare sector, where patient safety is of utmost importance, it is crucial for healthcare organizations to comprehend the potential risks associated with LLM usage.
When utilized to diagnose hypothetical patient cases, LLMs have exhibited accuracy at a level comparable to that of a third- or fourth-year medical student, albeit not reaching a professional's proficiency. 
Despite this high level of performance, LLMs have also been known to generate false information, invent sources, commit logical errors, and provide answers that are inappropriate or unethical. 
The integration of KGs would definitely enhance the capabilities of LLMs given the possibility of injecting domain-specific knowledge able to mitigate the issues mentioned above.
Avoiding hallucinations and preserving ethics are definitely the two major aspects to which LLMs+KG may contribute in a significant manner.

\textbf{Privacy}.
One major concern with LLMs is that employing any third-party application necessitates the transmission of data to that party. 
When data, including protected health information (PHI), is managed by a covered entity like a hospital, it becomes subject to the regulations of the jurisdiction where the entity is located (e.g., GDPR). 
Furthermore, by sending PHI to additional third parties, organizations lose control over how that data will be handled. 
For instance, healthcare organizations are unable to determine the exact storage location of their data, whether it will be mixed with data from other organizations and utilized to train future language models, or what security measures are in place to safeguard the data. 
Healthcare organizations employing LLMs must recognize that their data are potentially more susceptible to breaches or misuse.
The role of preserving private information may be played by KGs through the modeling of axioms defining which data may be shared and with who and how personal knowledge may be anonymized in order to be transmitted to possible external systems.

An alternative approach, prioritizing privacy, involves running an open-source LLM within the infrastructure of a healthcare organization. 
This way, it would be possible to directly work on the parameters of LLMs given the possibility of having control of the injected knowledge and, at the same time, ensuring that data are never shared with a third party. 
However, these open-source models are currently not as advanced or extensively trained as the more popular commercial systems (e.g., ChatGPT). 
Additionally, the effort to create the knowledge resources to inject and the expertise required to program and maintain an open-source LLM may not be readily available to many healthcare organizations.

The two risks described above lead to likewise challenges that must be tackled in order to make a significant step toward the adoption of these models within the clinical practice.

\subsubsection{Domain Specific Content Search 
}

Recently we have witnessed the success of models like GPT-4~\cite{openai2023gpt4} in a multitude of NLP applications that involve multiple modalities and domain specific adaptations.
While LLMs are often treated as generative models, they can be easily adopted in search and reasoning tasks in many tools and pipelines~\footnote{E.g., \url{https://haystack.deepset.ai/}}.
Nonetheless, as seen in the past all these specialized domains can better benefit from the inclusion of symbolic knowledge in machine methods~\cite{Alshahrani:NuroSym17}.
Here we focus on two distinct applications: (1) semantic image and video search and (2) technical document understanding.

Recently, many methods, primarily based on deep learning models such as CLIP~\cite{CLIP} and BLIP2~\cite{BLIP2}, achieved state-of-the-art performance on image retrieval tasks. 
These multi-modal models jointly learn vector embeddings for images and text, such that the embedding of the image should be close to that of the text that describes that image. 
Nonetheless, we have also seen increased  interest in obtaining a more symbolic representation of the contents of an image~\cite{VisualComet20,krishna2017visual}.
Datasets like  Visual Genome~\cite{krishna2017visual} annotate images with \emph{scene graphs}.
A scene graph is a small KG that describes, with a structured formal graphical representation, the contents of an image in terms of objects (people, animals, items) as nodes connected via pairwise relationships (e.g., actions or positioning relationships) as edges.
Therefore, Multimodal LLM can be trained to reason and exploit this additional representation offering an advanced ability to understand the contents of an image (or a video).
Importantly, a scene graph node and edge can be annotated (grounded) with features and positions from the image (e.g., relative size).
This can be exploited in applications like image and video search and question answering.
When entities in a scene graph are connected to a background taxonomy or KG, then questions that require forms of abstraction and computation, e.g., \emph{What is the genus of the largest animal in the picture?} \emph{What is the average price of the car in the picture?} \emph{Retrieve images depicting kids wearing vegan friendly items of clothing in the catalog}.

On the other hand, the digitalization of domain specific documents, e.g., especially contracts, is enabling in-depth applications of machine intelligence to help humans more effectively perform time-consuming tasks. 
Among these, contract review costs humans substantial time, money, and attention (many law firms spend approximately 50\% of their time reviewing contracts, costing hundreds of thousands of dollars)~\cite{CUAD:21}. 
The Contract Understanding Atticus Dataset (CUAD) is a new dataset for legal contract review~\cite{CUAD:21}. 
CUAD was created with legal experts and consisted of over 13,000 annotations.
Recent advancements in generic pre-trained language models showed their power in some text-understanding tasks~\cite{wu2023bloomberggpt,LLMMed:Yang2022}. 
Therefore, we have seen different domain specific applications of NLP and LLM methods~\cite{wu2023bloomberggpt,LLMMed:Yang2022}.
Yet, they do not have access to all relevant knowledge and are ill-suited for certain calculations~\cite{karpas2022mrkl}. 
This can be solved by exploiting external domain specific symbolic information, e.g., domain specific knowledge graphs~\cite{noy2019industry,lissandrini2015unleashing}, and by adding symbolic and reasoning capabilities to the LLMs~\cite{karpas2022mrkl}.
This promising direction will lead to extending current state-of-the-art neuro-symbolic methods to combine the advantages of a graph representation when extracting symbolic knowledge from complex documents, the ability to understand long-form unstructured texts of LLMs, and the good properties of domain-specific code synthesis of LLMs to address advanced retrieval and question answering use-cases, e.g., \emph{How many weeks are left before the expiration of this contract?} \emph{What are the conflicts between this liability clause and previous contracts my company signed?} \emph{What is the risk of side-effects of this treatment for a patient with this given health record?} \emph{Get the average yearly yield and spread for competitor companies of ACorp and their recent acquisitions in the EU market}.






\section{Outlook}
In conclusion, the recent advances on large language models (LLMs) mark an important inflection point for knowledge graph (KG) research. While important questions on the ability to combine their strengths remain open, these offer exciting opportunities for future research. The community is already rapidly adapting their research focus, with novel forums like the KBC-LM workshop \cite{kbc-lm-workshop} and the LM-KBC challenge \cite{singhania2022lm} arising, and resources massively shifting towards hybrid approaches to knowledge extraction, consolidation, and usage. We give out the following recommendations:
\begin{enumerate}
    \item \textit{Don't throw out the KG with the paradigm shift}: For a range of reliability or safety-critical applications, structured knowledge remains indispensible, and we have outlined many ways in which KGs and LLMs can fertilize each other. KGs are here to stay, do not just ditch them out of fashion.
    \item \textit{Murder your (pipeline) darlings}: LLMs have substantially advanced many tasks in the KG and ontology construction pipeline, and even made some tasks obsolete. Take critical care in examining even the most established pipeline components, and compare them continuously with the LLM-based state of the art.
    \item \textit{Stay curious, stay critical}: LLMs are arguably the most impressive artifact of AI research of the past years. Nonetheless, there exist a magnitude of exaggerated claims and expectations in the public as well as in the research literature, and one should retain a healthy dose of critical reflection. In particular, a fundamental fix to the so-called problem of hallucinations is not in sight.
    \item \textit{The past is over, let's begin the new journey}: The advances triggered by LLMs have uprooted the field in an unprecedented manner, and enable to enter the field with significant shortcuts. There is no better time to start anew in fields related to Knowledge Computing, than now.
\end{enumerate}
Although the direction of the present transformation is widely open, as researchers continue to explore the potentials and challenges of hybrid approaches, we can expect to see new breakthroughs in the representation and processing of knowledge, with far-reaching implications for fields ranging from Knowledge Computing to NLP, AI, and beyond.

\bibliographystyle{plain}
\bibliography{bibs/sample-base}

\end{document}